\documentclass[onecolumn,draftcls]{IEEEtran}
\usepackage{cite}
\usepackage[pdftex]{graphicx}
\usepackage{caption}
\usepackage{amsmath} 
\usepackage{amsfonts} 
\usepackage{xcolor}
\usepackage{amssymb}

\begin{document}
\title{Thyroid Cancer Malignancy Prediction From Whole Slide Cytopathology Images}

\author{David~Dov,
        Shahar~Kovalsky,
        Jonathan~Cohen,
        Danielle~Range,
        Ricardo~Henao,
        and~Lawrence~Carin

\thanks{D. David, R. Henao and L. Carin are with the Department of Electrical and Computer Engineering, Duke University, Durham, NC 27708, USA (e-mail: david.dov@duke.edu; ricardo.henao@duke.edu; lcarin@duke.edu).}

\thanks{S. Kovalsky is with the Department of  Mathematics, Duke University, Durham, NC 27708, USA (e-mail: shaharko@math.duke.edu).}

\thanks{J. Cohen is with the Department of Surgery, Duke University Medical Center, Durham, NC 27710, USA (e-mail: jonathan.m.cohen@duke.edu).}

\thanks{D. Elliott Range is with the Department of Pathology, Duke University Medical Center, Durham, NC 27710, USA (e-mail: danielle.range@duke.edu).}

}
\maketitle

\begin{abstract}
  We consider preoperative prediction of thyroid cancer based on ultra-high-resolution whole-slide cytopathology images. Inspired by how human experts perform diagnosis, our approach first identifies and classifies diagnostic image regions containing informative thyroid cells, which only comprise a tiny fraction of the entire image. These local estimates are then aggregated into a single prediction of thyroid malignancy. Several unique characteristics of thyroid cytopathology guide our deep-learning-based approach. While our method is closely related to multiple-instance learning, it deviates from these methods by using a supervised procedure to extract diagnostically relevant regions. Moreover, we propose to simultaneously predict thyroid malignancy, as well as a diagnostic score assigned by a human expert, which further allows us to devise an improved training strategy. Experimental results show that the proposed algorithm achieves performance comparable to human experts, and demonstrate the potential of using the algorithm for screening and as an assistive tool for the improved diagnosis of indeterminate cases. 
\end{abstract}

\section{Introduction}
Thyroid cancer is one of the most common cancers worldwide; it is predicted to be the third most common cancer in women in the US by 2019 \cite{aschebrook2013clinical}.  Among the various diagnostic tests available, the analysis of a thyroid biopsy is the most important step in the \emph{preoperative} diagnosis of thyroid cancer \cite{popoveniuc2012thyroid}. This  analysis is performed by an expert pathologist who manually examines the biopsy tissue under a microscope in order to estimate the risk of malignancy. In this work we establish a dataset of \emph{whole-slide} digital thyroid cytology images and propose a deep-learning-based algorithm for computational preoperative prediction of thyroid malignancy.

A thyroid nodule (mass) is typically detected by palpation or found incidentally in ultrasound imaging.  Once discovered, preoperative evaluation includes ultrasound characterization followed by biopsy for nodules larger than 1 cm.  Fine needle aspiration biopsy (FNAB) is performed using a thin, hollow needle which is inserted into the thyroid nodule to extract cells which are then smeared onto a glass microscope slide and stained. The glass slide is then examined by an expert in \emph{cytopathology}, a discipline of pathology that uses individual cells and cell groups for diagnosis, in contrast to histopathology which uses whole tissues sections. The cytopathologist looks at the stained slide (smear) under a light microscope and assesses the risk of thyroid malignancy according to various cellular features, such as cell size, nuclear size, and architectural features of cell groups; these features are used to map the slide to a ``score,'' using the Bethesda System for the Reporting of Thyroid Cytopathology (TBS) \cite{cibas2009bethesda}. TBS is the universally accepted reporting system for thyroid FNAB diagnosis and comprises six categories/scores: TBS categories $2$ and $6$ classify the biopsy as benign and malignant, respectively, and are both associated with clear treatment guidelines; benign lesions undergo surveillance and malignant ones undergo surgery. TBS $3$, $4$, and $5$ are considered indeterminate categories associated with an \emph{increased} risk of malignancy. TBS $1$ is of non-diagnostic quality, and is not considered in this study.

In most healthcare systems, however, cytopathology slides are not routinely digitized. Therefore, as an integral part of this research, we have established a dataset of $908$ samples. Each sample consists of: $i$) high-resolution digital cytopathology scan, with typical resolution of $150,000\times 100,000$ pixels; $ii$) \emph{preoperative} TBS category assigned by a cytopathologist, as recorded in the medical files, and $iii$) \emph{postoperative} histopathology diagnosis. The latter is the gold standard for determining malignancy, and is considered the \emph{ground truth} for this study.  

We take a machine learning approach for predicting thyroid malignancy from whole-slide cytopathology slides. The use of machine learning for pathology, as well as related problems in medical imaging, has been receiving attention in recent years, as detailed in Subsection~\ref{subsec:related_work}. Nonetheless, the problem of \emph{fully automated} prediction of malignancy from \emph{whole-slide} cytopathology slides remains largely unaddressed. The use of  entire slides poses significant challenges due to the huge, computationally prohibitive, size of digital scans. The typical size of a single slide is tens of gigabytes, and thus cannot be fed in its entirety into current GPUs, as it exceeds their memory limitations. Moreover, only a small and scattered fraction of a slide contains follicular (thyroid) cells relevant for prediction; whereas the majority of other regions are irrelevant (\emph{e.g.}, red blood cells) and are to be considered background. Therefore, a key element of the proposed approach concerns how to split the digital scan into multiple diagnostically relevant image regions and, in turn, how to use these to compute a global, slide-level prediction of thyroid malignancy. 

To that end, we take an approach inspired by the work of cytopathologists, who initially overview the slides at low magnification to identify areas of diagnostic interest, which are then examined at higher magnification to assess the cell features. Specifically, we use a small set of sample patches, localized and annotated by an expert cytopathologist, to train a supervised model for distinguishing groups of follicular cells from irrelevant, background regions. In turn, image regions containing follicular cells are used to predict thyroid malignancy.

Our approach further exploits the observation that the preoperative TBS category, determined by an expert pathologist, is a monotone and a consistent proxy for the probability of malignancy: The higher the TBS category, the higher the probability of malignancy. This is supported by the observation that nearly $100\%$ of the cases categorized as TBS 2 and 6 are indeed benign and malignant, respectively. Furthermore, the TBS categories assigned by different experts are highly unlikely to differ by more than 1 or 2 TBS categories \cite{jing2012group,pathak2014implementation}. Similar behavior is essential for the clinical applicability of a machine learning approach. Towards this end, we propose to jointly predict both the TBS category as well as the probability of malignancy. We compute these predictions from a single output, which in turn implicitly enforces monotonicity and consistency.

\paragraph{Technical Significance}
We address the task of predicting thyroid malignancy from cytopathology whole-slide images. We propose a prediction algorithm based on a cascade of two convolutional neural networks. The first network identifies image regions containing groups of follicular cells, while learning to ignore irrelevant background findings such as blood. The second network looks at follicular groups to make a global, slide-level prediction of malignancy. We propose a training strategy based on unique characteristics of thyroid cells, and on the simultaneous prediction of thyroid malignancy and the preoperative TBS category, from a single output of the second network. The prediction of the TBS category further acts as a regularizer that improves our predictor. 

\paragraph{Clinical Relevance}
The proposed algorithm was tested on a dataset of $109$ whole-slide images, never seen during the training procedure. As shown in our experiments, our predictions of thyroid malignancy are comparable to those of $3$ cytopathology experts (who, for this research, analyzed the same data as the algorithm), as well as the pathologists on record (who originally evaluated each case at the time the FNAB was performed). In particular, we observe that all cases for which the predicted TBS category is $2$ and $6$ were indeed benign and malignant, respectively. Moreover, these include cases which were assigned an indeterminate TBS category $3$ to $5$ in the medical records. Our results demonstrate the clinical potential of our approach, which could be used as a screening tool to streamline the cytopathological evaluation of FNABs as well as possibly resolve indeterminate diagnoses and improve treatment.

\section{Background}
\subsection{Problem Formulation}
We begin by dividing each image into a set of $M$ small image regions (patches). We denote the $m$-th patch by $\mathbf{x}_{m}\in\mathbb{R}^{w\times h\times3}$, with $w$ and $h$ being the width and the height of the patch, respectively. Given these image regions, the goal is to predict thyroid malignancy, denoted by $Y\in\left\{ 0,1\right\} $ and referred to as the \emph{global} label of the image, where $0$ and $1$ correspond to benign and malignant cases, respectively. In addition, we propose to predict the TBS category assigned to the slide by a pathologist, which we denote by $S\in\left\{2, 3, 4, 5, 6\right\}$. In Section \ref{sec:methods}, we propose to train a neural network to simultaneously predict $Y$ and $S$ so as to provide more accurate and reliable predictions of thyroid malignancy.

We further consider a \emph{local} label $y_m\in\left\{ 0,1\right\}$ associated with the $m$-th patch, taking a value $y_m=1$ if the patch $\mathbf{x}_{m}$ contains a group of follicular cells, and zero otherwise.  These local labels are used to train a convolutional neural network for the identification of patches containing follicular groups, while ignoring other image regions containing background. In turn, regions selected by this neural net are used, in a second stage, to predict thyroid malignancy.

\subsection{Related work}\label{subsec:related_work}
\paragraph{Machine learning in medical imaging.}
This study is related to a rapidly growing body of work on the use of machine learning, and in particular deep neural networks, in medical imaging \cite{litjens2017survey}. Such technology has been used in the classification of skin cancer \cite{esteva2017dermatologist}, the detection of diabetic retinopathy \cite{gulshan2016development}, in histopathology \cite{litjens2016deep,djuric2017precision,sirinukunwattana2016locality} and cytopathology \cite{pouliakis2016artificial}. The use of machine learning for the prediction of thyroid malignancy from ultrasound images and from histopathology tissue sections has been studied in \cite{liu2017classification,chi2017thyroid,ma2017ultrasound,ma2017pre,li2018improved,song2018multi,ozolek2014accurate}. Most related to the clinical question we address is work concerned with thyroid cytopathology \cite{daskalakis2008design,varlatzidou2011cascaded, gopinath2013computer,kim2016deep, sanyal2018artificial, gilshtein2017computerized, savala2018artificial}. These however consider algorithms tested on a small number of individual cells or ``zoomed-in'' regions manually selected by an expert cytopathologist. However, the challenge of predicting malignancy from whole-slide cytopathology images remains largely unaddressed. 

\paragraph{Multi instance learning.}
The setup of using multiple instances (image regions) grouped into a single bag (digital scan) associated with a global label is typically referred to in the literature as multiple instance learning (MIL). Typical MIL approaches aggregate separate predictions generated from each instance into a global decision, using pooling functions such as noisy-or \cite{zhang2006multiple} and noisy-and \cite{kraus2016classifying}. In a recent MIL approach \cite{ilse2018attention}, the instances are represented by features learned by a neural network; the bag-level decision is obtained from a weighted average of the features, using an attention mechanism. Example applications of MIL approaches include classification and segmentation of fluorescence microscopy imagery of breast cancer \cite{kraus2016classifying}, prediction of breast and colon cancer from histopathology slide scans \cite{ilse2018attention}, and breast cancer prediction from mammograms \cite{quellec2017multiple}.

Common to these approaches is the underlying assumption that only relevant instances are manifested in the prediction and that the effect of irrelevant instances is reduced by the pooling or weighting components. In our case, groups of follicular cells are most relevant for the prediction of malignancy; their characteristics, such as architecture, size and texture, are the main cue according to which cytopathologists determine malignancy and assign the TBS category. However, follicular cells constitute merely a tiny fraction of the digital scan, which led in our experiments to poor performance of classic MIL approaches. 

\paragraph{Detection of diagnostic regions.}
The detection of diagnostically relevant regions, that contain follicular cells, is a key component of our approach. The detection of regions of interest is widely studied in the context of object detection \cite{uijlings2013selective,girshick2014rich,girshick2015fast,ren2017faster}. However, our task differs from classic object detection, which is typically focused on the accurate detection of the boundaries of the objects (instances) and their individual classification into different classes. In contrast, we are ultimately interested in the prediction of one global label for the entire scan.

\paragraph{Ordinal regression.}
Our approach predicts the TBS category in conjunction to malignancy. Predicting the TBS category can be viewed as a multi-class classification problem, where the different classes have a meaningful increasing order; namely, wrong predictions are considerably worse the more they differ from the TBS category assigned by an expert cytopathologist. This problem is often referred to in the literature as ordinal regression \cite{gutierrez2016ordinal}. Particularly relevant to this paper is a class of methods termed cumulative link models, in which a single one-dimensional real output is compared to a set of thresholds for the classification \cite{mccullagh1980regression, agresti2003categorical,dorado2012ordinal}.

\section{Cohort}
In standard practice, cytopathology slides are assessed using an optical microscope and are \emph{not} routinely digitized. Therefore, as an integral part of this research, we have established a dataset of thyroid cytopathology whole slide images.

\paragraph{Design of cohort.}
The data collection for this research was approved by and conducted in compliance with the Institutional Review Board. We searched the institutional databases for all thyroid cytopathology (pre-operative) fine needle aspiration biopsies (FNAB) with a subsequent thyroidectomy surgery from June 2008 through June 2017. The ground truth postoperative pathology and preoperative TBS category corresponding to each slide were manually extracted from the medical record and pathology reports.  For patients with more than one thyroid nodule or multiple FNABs, the medical record was reviewed to ensure correlation between any individual nodule and the preoperative and postoperative diagnoses;  cases for which this correlation could not be established were excluded from the study. The final cohort includes $908$ cases for which both a preoperative cytopathology slide as well as a postoperative determination of malignancy were available. For each case, we selected a representative, single alcohol-fixed, Papanicolaou-stained FNAB slide. Each slide was de-identified (all patient health information removed) and assigned a random study number.

\paragraph{Acquisition.}
Cytopathology slides were digitized using an Aperio AT2 scanner by Leica Biosystems. Scanning was performed with a 40$\times$ objective lens, resulting in an image wherein a pixel corresponds to 0.25$\mu$m. The area to be scanned was manually adjusted by a technician so as to avoid scanning empty areas and to reduce scanning time; consequently, the size of scanned images vary, with a typical resolution of $150,000\times 100,000$ pixels. Scanning was performed at an automatically set focal plane (\emph{i.e.}, auto-focus) as well as at 4 focal planes above and 4 focal planes below, equally spaced by 1$\mu$m; thus resulting in a focal stack (termed \emph{z-stack} in digital pathology) of $9$ images, wherein the central image corresponds to the auto-focus. Overall, we acquired a dataset of over $20$ terabytes of images. To make the runtime of our experiments reasonable, in this work we only use a single focal plane (the central, auto-focused image) which we further downscale by a factor of $4$. Our experiments demonstrate that these settings are sufficient to produce thyroid malignancy predictions approaching human level. Utilization of all Z-stacks and full resolution data will be done in a future study.

\paragraph{Test and train data.}
We designated a subset of $109$ slides to be exclusively used as a test set. These slides were selected in chronological order, going back from June 2016. The slides were reviewed by a an expert cytopathologist who examined the digital scans at multiple focal planes using the Aperio ImageScope software  (rather than the glass slide) and assigned a TBS category. Technically-inadequate slides, for which the cytopathologist could not assign a TBS category due to the slide being uninformative (containing less that $6$ groups of follicular cells) or the scan being severely out-of-focus, were excluded from the test set. Each glass slide in the designated test set was further reviewed by $2$ additional expert cytopathologists, each of whom assigned a TBS category according to their assessment. In their reviews, all $3$ experts only had access to the de-identified slide itself, and any additional information ($e.g.$, the BTS category assigned in the medical record) was omitted. The test set slides were not seen by the algorithm during training.

The remaining $799$ slides, along with their ground truth labels as described above, were used for training the proposed algorithm. The scan of these slides were not individually reviewed by a cytopathologist as in the case of the test set, thus the training data may include slides that are uninformative as well as out-of-focus scans. A subset of $142$ of these slides were further locally-annotated by an expert cytopathologist, who reviewed the images and manually annotated an overall of $5461$ diagnostically indicative image regions that contain follicular groups.

\section{Methods\label{sec:methods}}

\paragraph{Detecting groups of follicular cells}
We consider groups of follicular cells as the main cue to predict thyroid malignancy. Benign and malignant groups differ from each other in the shape and the architecture of the group, size and texture nuclear color, and tone of the cells and their number.

\begin{figure}[t!]
	\centering
	\includegraphics[width=.9\columnwidth,trim={18mm 25mm 18mm 25mm},clip]{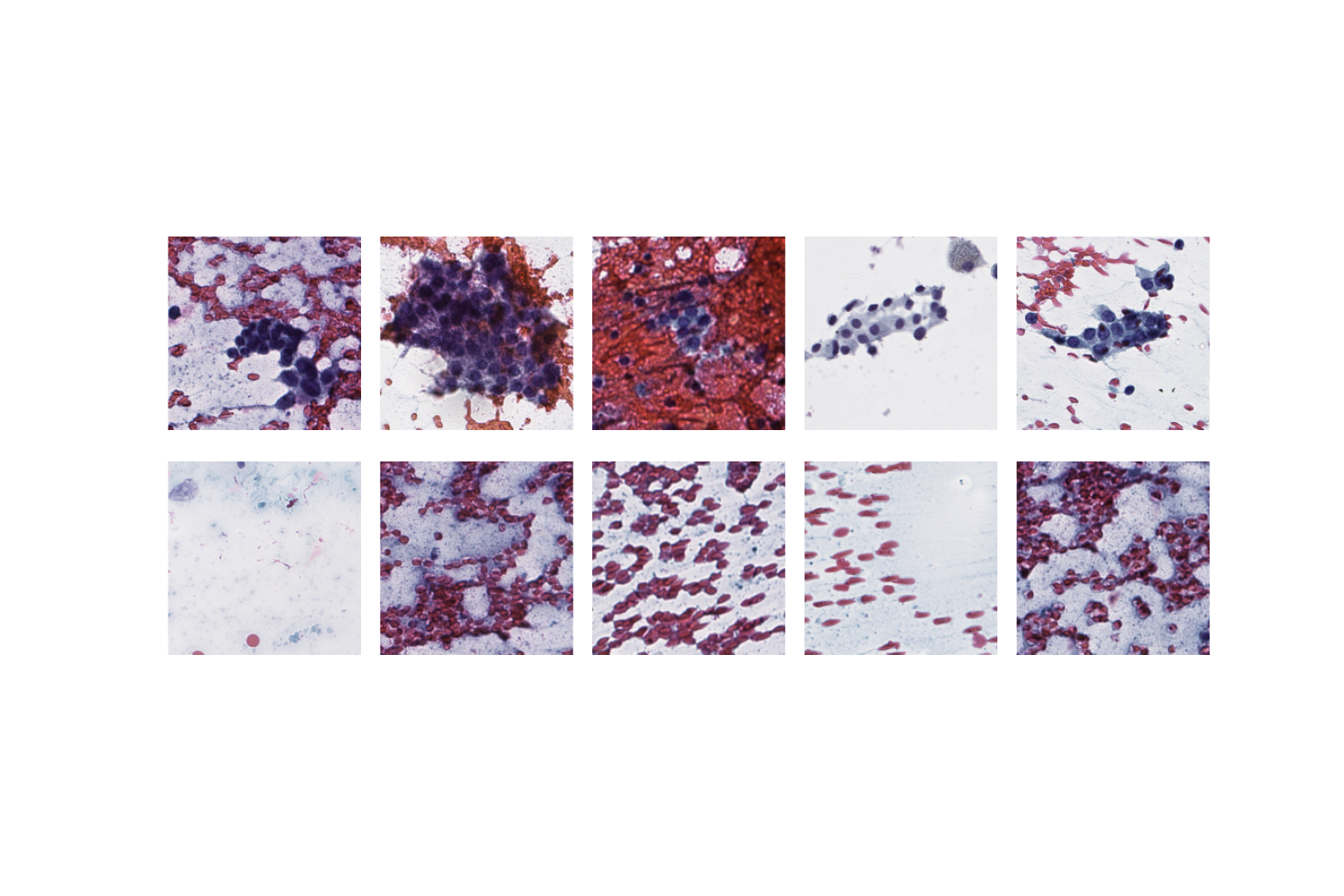}
	\caption{(Top) Examples of image regions containing groups of follicular cells. (Bottom) Examples of background image regions.}
	\label{fig:ROIS}
\end{figure}

Despite being the informative cells, follicular cells constitute only a tiny fraction of the scan, whereas the vast majority of image regions contain mainly irrelevant blood cells, and are considered background. Examples of image regions containing follicular groups, as well as regions containing background are presented in Figure~\ref{fig:ROIS}.

To distinguish between the two groups, we use the local labels $\{ y_{m}\}_{m=1}^M$ and train a convolutional neural network to classify the patches $\{\mathbf{x}_{m}\}_{m=1}^M$. We use an architecture based on VGG11 \cite{simonyan2014very}, which is considered small and fast-converging; details of the architecture are summarized in Appendix A.
We note that the challenge in training the network properly to identify follicular groups is in having a sufficient amount of labeled patches. However, the collection of such data would require manual examination of the scans, which is a prohibitively time consuming procedure to be done by an expert cytopathologist. Therefore, we focus the labeling efforts on only marking \emph{positive} examples $(y_m=1)$ of follicular groups. We further observe in our experiments that since only a tiny fraction of the scan contains follicular groups, randomly sampled patches typically contain background. Therefore, we obtain negative examples $(y_m=0)$ by sampling patches uniformly at random from the whole slide, under the assumption they are likely to contain background. Despite the uncertainty in the accuracy of the negative examples, we observed in our experiments that the network successfully distinguishes between follicular groups and the background, and, in turn, thyroid malignancy is successfully predicted from the selected regions, as we show in our experiments.

We apply the network for the identification of follicular groups to the full scans, and select a subset of $\tilde{M}\ll M$  image regions providing the highest predictions, recalling that $M$ is the number of patches in the scan. We use $\tilde{M}=1000$ to train the malignancy prediction network; we found this value to provide a good balance between using patches with sufficiently high prediction value of being follicular groups and having a sufficient amount of patches to train the malignancy predictor. During testing, we use $\tilde{M}=100$, a value selected using a validation set, that slightly improves the performance.

\paragraph{Predicting thyroid malignancy from multiple image regions}
Using the $\tilde{M}$ patches containing follicular groups, we propose a simple, yet effective, procedure for predicting thyroid malignancy. We separately feed each patch into a second neural network, which we denote by $g\left(\cdot\right) \in \mathbb{R}$. The architecture of $g\left(\cdot\right)$ is similar to the first neural network and is trained with the same hyper-parameters  (see Appendix A for details). We refer to the outputs of $g\left(\cdot\right)$ as patch-level (local) decisions, which are then averaged into a single slide-level prediction:
\begin{equation}
	\hat{Y}=\frac{1}{\tilde{M}}\sum_{m}g\left(\mathbf{x}_{m}\right) \triangleq \bar{g}.
    \label{eq:proposed_mil}
\end{equation}
We note that obtaining a global prediction $\hat{Y}$ in \eqref{eq:proposed_mil} from a set of multiple image regions (instances) is typically referred in the literature as multiple instance learning. Classical MIL approaches such as \cite{kraus2016classifying,ilse2018attention} focus on implicitly identifying the informative instances and assigning high weights in their aggregation. In contrast, we use a supervised procedure for the identification of the informative regions, which we found provides significantly better predictions.

\paragraph{Training strategy} 
We propose to train $g\left(\cdot\right)$ to predict the global label using the the binary cross entropy loss (BCE):
\begin{equation} 	Y\log\left[\sigma\left(g\left(\mathbf{x}_{m}\right)\right)\right]+\left(1-Y\right)\log\left[1-\sigma\left(g\left(\mathbf{x}_{m}\right)\right)\right],
    \label{eq:loss_pth}
\end{equation}
where $\sigma\left( \cdot \right)$ is the sigmoid function. Equation \eqref{eq:loss_pth} suggests to train the network by separately predicting the global label $Y$ from each single patch $\mathbf{x}_m$ rather than using multiple patches by replacing $g\left(\mathbf{x}_{m}\right)$ with the global prediction $\bar{g}$. The latter is common practice in MIL methods such as those presented in \cite{kraus2016classifying,ilse2018attention}. These MIL approaches assume that a global label is positive if at least one of the instances (patches) is positive. Instead, \eqref{eq:loss_pth} is based on the observation that in a malignant slide \textit{all} follicular groups are malignant, whereas in a benign slide \textit{all} groups are benign. This observation is clinically supported by the FNAB procedure, since follicular groups are extracted via a fine needle from a single homogeneous lesion in the suspicious thyroid nodule. Indeed, we found in our experiments that the training strategy in \eqref{eq:loss_pth} further improves the predictions.

\paragraph{Simultaneous prediction of malignancy and Bethesda category}  We propose to predict the TBS category by comparing the output of the network to  threshold values $\left\{\tau_l\right\}_{l=1}^6 $, $\tau_l\in \mathbb{R}$ , yielding:
\begin{equation*}
\hat{S}=\left\{ \begin{array}{cc}
l; & {\rm if}\,\tau_{l-1}<\bar{g}<\tau_{l}
\end{array}\right\},
\label{pred_y_bts}
\end{equation*}%
where $\hat{S}$ is the prediction of the TBS category taking an integer value between $2$ to $6$; the threshold values satisfy $\tau_1<\tau_2<\cdots<\tau_6$. The thresholds $\tau_2, \dots, \tau_5$ are learned along with the parameters of the network, and  $\tau_1\rightarrow -\infty$  and $\tau_6\rightarrow \infty$. 

The prediction of TBS category is based on the proportional odds ordinal regression model, presented in \cite{mccullagh1980regression, dorado2012ordinal}, often referred to as a cumulative link model. The core idea is to pose order on the predictions of $S$ by penalizing predictions the more they deviate from the true TBS category. This is obtained by predicting $P\left(S>l\right)$, \emph{i.e.}, the probability that the TBS is higher then a certain category from the output of the network:
\begin{equation}
	P\left(S>l\right) = \sigma\left(\bar{g}-\tau_{l}\right)
\end{equation}
Specifically, using BCE loss, we propose the following loss function:
\begin{equation}
	\sum_{l=2}^{5}S_{l}\log\left[\sigma\left(g\left(\mathbf{x}_m\right)-\tau_{l}\right)\right] 	+ \left(1-S_l\right)\log\left[1-\sigma\left(\left(\mathbf{x}_m\right)-\tau_{l}\right)\right],
	\label{eq:loss_bts}
\end{equation}
where $S_{l}\triangleq P\left(S>l\right)$. Namely, $S_l$ are labels used to train $4$ classifiers corresponding to $l\in\left( 2, 3, 4, 5\right)$, whose explicit relation to TBS categories is given in Table \ref{tab:ordinal_gr_truth}.  The use of $g\left(\cdot\right)$ in \eqref{eq:loss_bts} instead of $\bar{g}$ follows from the training strategy discussed above. For more detailed description of the ordinal regression framework we refer the reader to \cite{dorado2012ordinal}. 

The total loss function we use for simultaneously predicting thyroid malignancy and the TBS category  is given by the sum of \eqref{eq:loss_pth} and \eqref{eq:loss_bts}.  We interpret the joint estimation of malignancy and TBS as a cross-regularization scheme: Consider for example a case assigned (by a pathologist) with TBS 5 which eventually turned out benign; in such a case, the term of the loss corresponding to malignancy encourages low prediction values, whereas the term corresponding to TBS penalizes them.  On the one hand, the network is not restricted to the way a human implicitly predicts malignancy via the Bethesda system, as would be the case if it was trained to solely predict TBS. On the other hand, the network inherits properties of the Bethesda system, which we consider to be a good and reliable proxy for the risk of malignancy. In particular, the output of the network $\bar{g}$ better resembles the probability of malignancy such that the higher the prediction value the higher is the probability of malignancy. 

\begin{table}[htbp]
    \centering

  \begin{tabular}{|c|c||c|c|c|c|}
    \cline{3-6} 
    \multicolumn{1}{c}{} & \multicolumn{1}{c|}{} & $S_2$ & $S_3$ & $S_4$ & $S_5$\tabularnewline
    \hline 
    \hline 
     & $2$ & $0$ & $0$ & $0$ & $0$\tabularnewline
    \cline{2-6} 
    TBS & $3$ & $1$ & $0$ & $0$ & $0$\tabularnewline
    \cline{2-6} 
    category & $4$ & $1$ & $1$ & $0$ & $0$\tabularnewline
    \cline{2-6} 
     & $5$ & $1$ & $1$ & $1$ & $0$\tabularnewline
    \cline{2-6} 
     & $6$ & $1$ & $1$ & $1$ & $1$\tabularnewline
    \hline 
  \end{tabular}
  \caption{Binary labels used in the proposed ordinal regression framework to predict the Bethesda score.}
    \label{tab:ordinal_gr_truth}
\end{table}

\section{Results}\label{sec:results} 
\begin{figure}[htbp]
\begin{centering}
\includegraphics[scale=0.9]{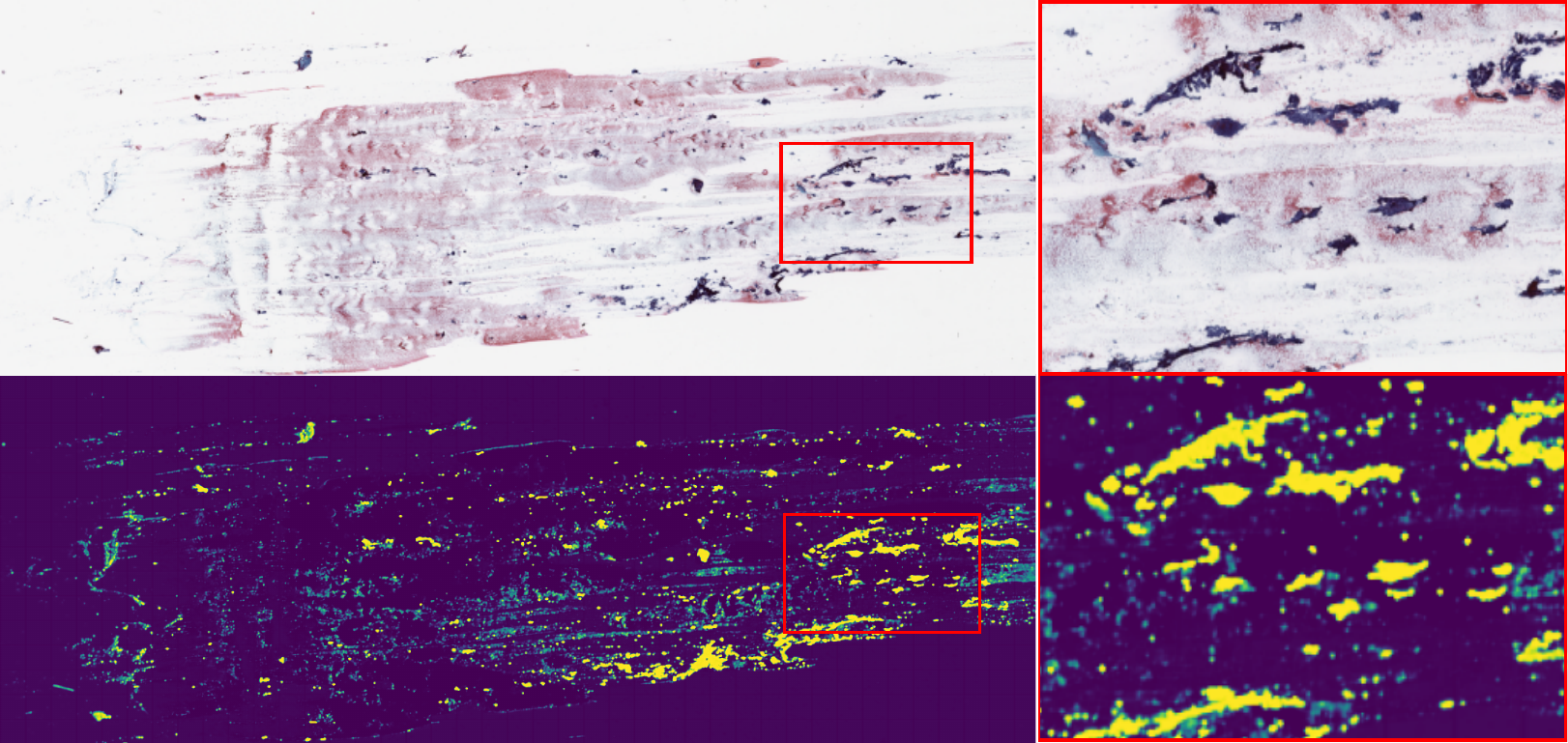}
\par\end{centering}
\caption{(Top left) Scan of a whole cytopathology slide. (Top right) Zoom in of the region marked by the red rectangle. (Bottom left) Heat map of the predictions of the first neural network. Bright regions correspond to image regions predicted to contain follicular groups. (Bottom right) Zoom in of the region marked by the red rectangle.}
\label{fig:full_slide_heat_map}
\end{figure}

\begin{figure}[htbp]
  \begin{centering}
  \includegraphics[viewport=90bp 30bp 350bp 260bp,clip, scale=1.7]{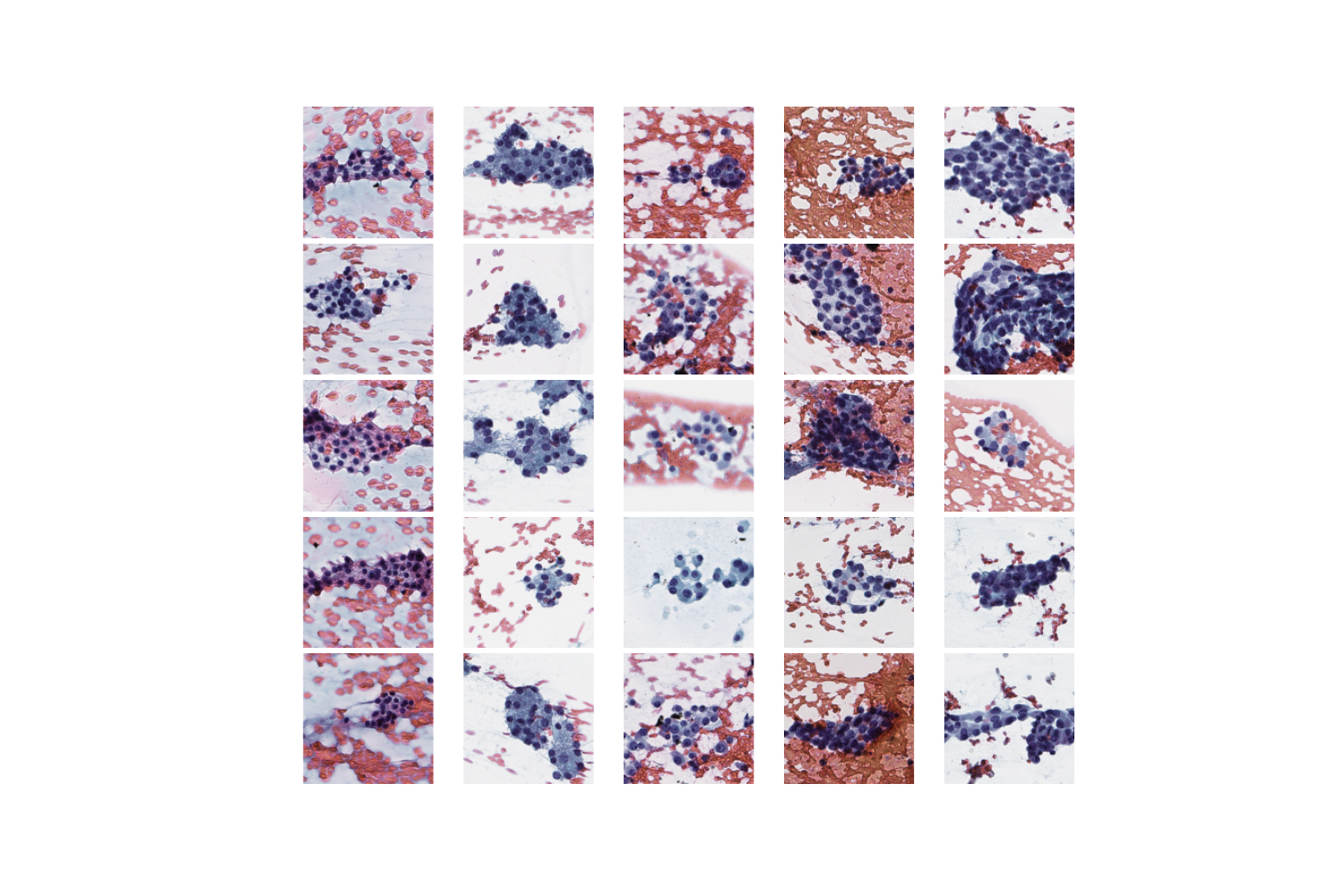}
  \par\end{centering}
  \centering{}
\caption{Examples of image regions containing follicular groups. The columns, from left to right, correspond to TBS $2-6$ cases.}
\label{fig:rois_bts}
\end{figure}

\paragraph{Identification of image regions with follicular groups}
Figure \ref{fig:full_slide_heat_map} shows a heat map illustrating how the first network identifies regions containing follicular groups of a representative scan. It can be seen that the vast majority of the patches contain background.
The bright regions in the heat map in Figure~\ref{fig:full_slide_heat_map} (bottom) correspond to the follicular groups seen in Figure~\ref{fig:full_slide_heat_map} (top). The high predictions indeed correspond to patches containing follicular groups, which we select for thyroid malignancy prediction. In Figure \ref{fig:rois_bts}, we present examples of the detected image regions containing follicular groups.

\begin{figure}[htbp]
\begin{centering}
\includegraphics[viewport=0bp 5bp 400bp 260bp,clip,scale=0.53]{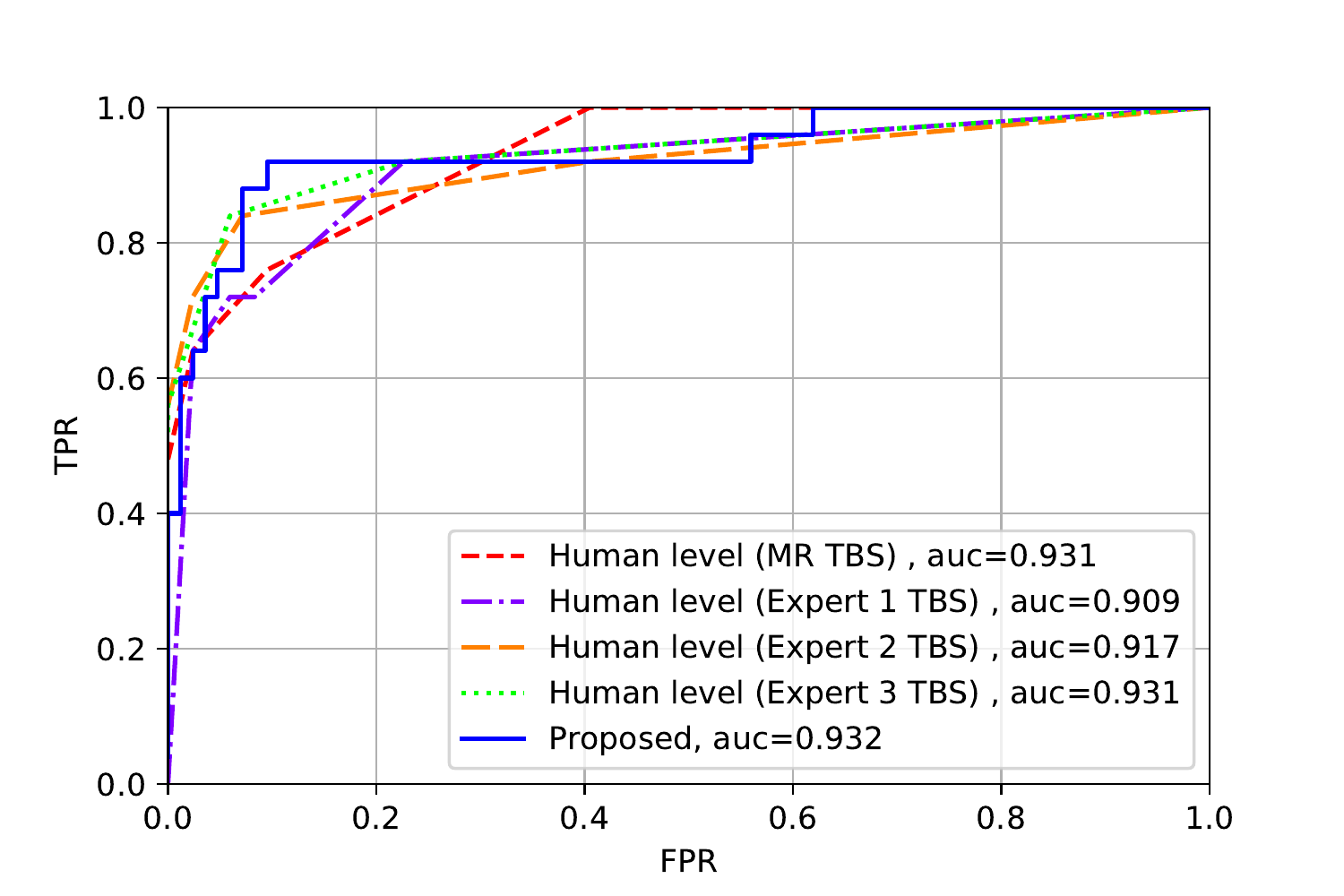}~\includegraphics[viewport=0bp 5bp 400bp 260bp,clip,scale=0.53]{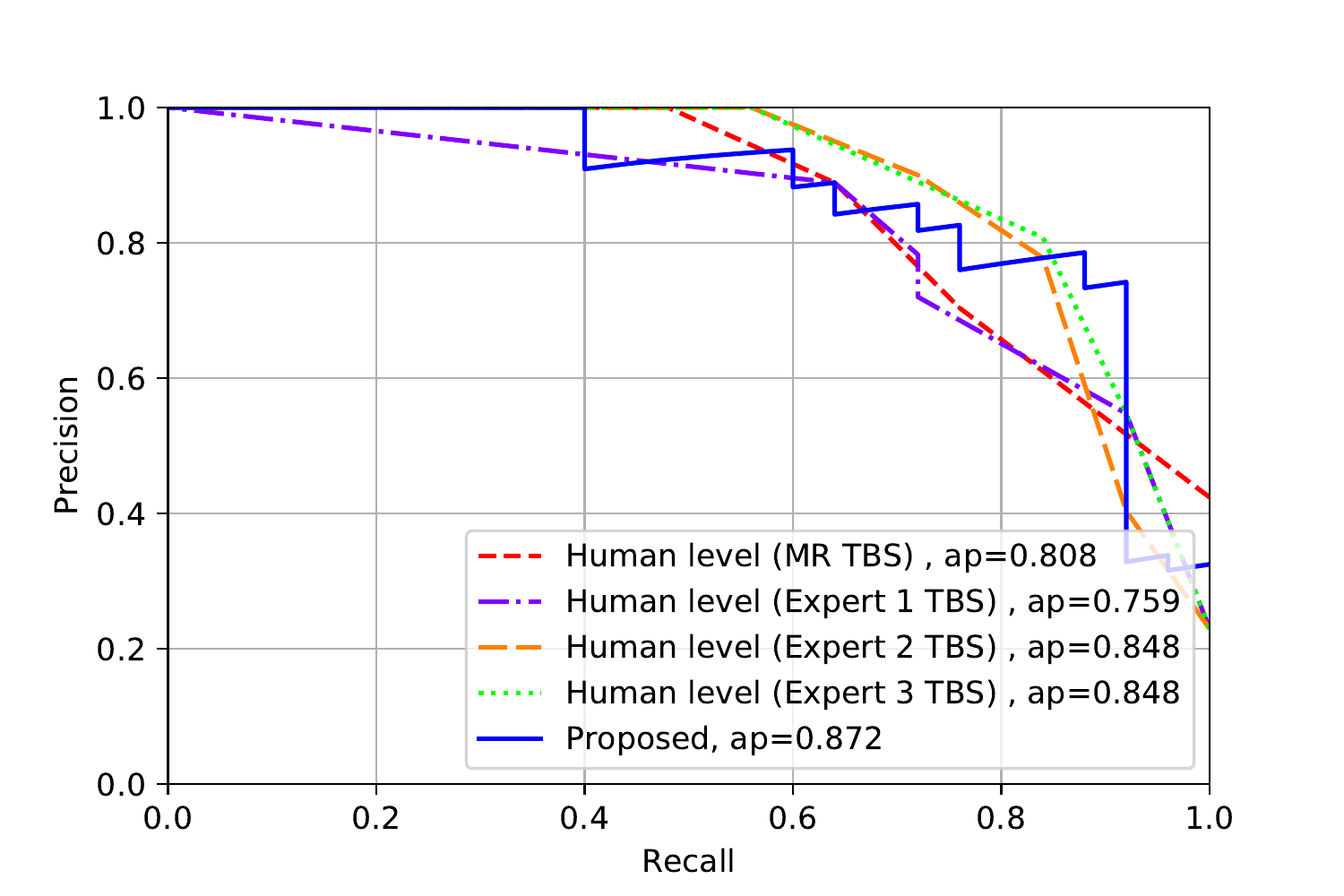}
\par\end{centering}
\centering{}
\caption{Comparing the performance of the proposed algorithm (blue curve) to human level thyroid malignancy predictions of three expert cytopathologists (purple, orange and green curves) and the pathologist on record (red curve, extracted from medical records). (Left) ROC and (Right) PR curves.}\vspace{-8pt}
\label{fig:roc}
\end{figure}

\paragraph{Thyroid malignancy prediction}
We consider the TBS category assigned by a pathologist as a ``human prediction of malignancy" where TBS $2$ to $6$ correspond to increasing probabilities of malignancy. In Figure~\ref{fig:roc}, we compare the performance of the proposed algorithm to the human predictions in the form of receiver operating characteristic (ROC) and precision-recall (PR) curves. In addition to the TBS category obtained from the medical record (MR TBS in the plot), the algorithm is compared to $3$ expert cytopathologists (Expert 1-3 TBS).  It can be seen in the plot that the proposed algorithm provides comparable AUC scores and improved AP scores compared to cytopathologists.

\begin{figure}[htbp]
	\begin{centering}
	\includegraphics[viewport=0bp 0bp 280bp 280bp,clip,scale=0.7]{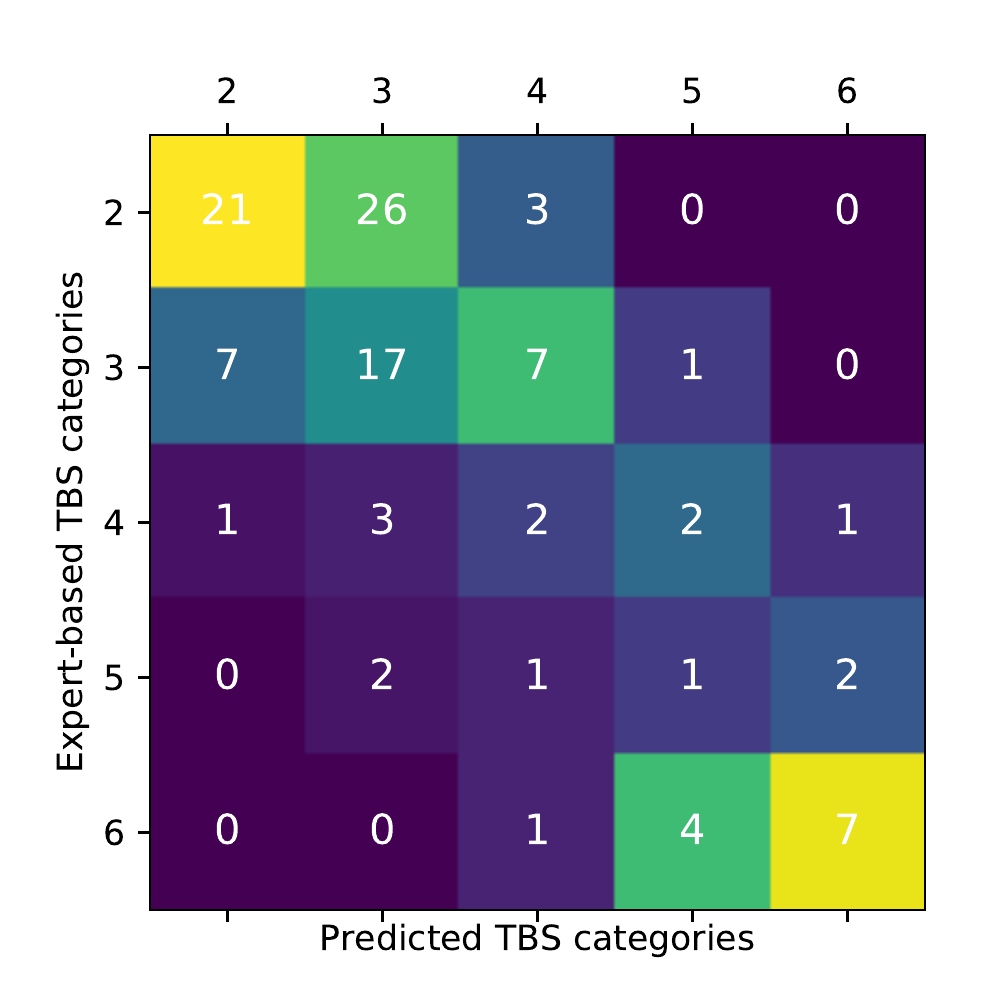}
	\par\end{centering}
	\centering{}
	\caption{Confusion matrix of predicted \emph{vs.} expert-based TBS categories. Colors correspond to a column normalized version of the confusion matrix.}
	\label{fig:tbs_vs_predicted}
\end{figure}

To gain further insight on the performance of the proposed algorithm, we also evaluate its TBS predictions, summarized in Figure~\ref{fig:tbs_vs_predicted}. We note that having high values on the main diagonal in the confusion matrix is not what we expect from the algorithm. Specifically, we do expect large number of TBS $2$ and $6$ predictions, when the TBS assigned by the pathologist is $2$ and $6$, respectively, since these decisions of pathologists has a high confidence of more than $97\%$ being benign and malignant, respectively. Indeed, high values are observed in the left-top and right-bottom cells in Figure~\ref{fig:tbs_vs_predicted}. On the other hand, original TBS $3, 4$ and $5$ are the indeterminate cases. In these cases, the algorithm is not trained to predict the exact TBS. According to the loss function which is the sum of \eqref{eq:loss_pth} and \eqref{eq:loss_bts}, it is trained to provide higher values in malignant cases and lower values in the benign ones compared to the original TBS. Therefore, reliable predictions of the algorithm correspond to a block diagonal structure, which is indeed observed in Figure~\ref{fig:tbs_vs_predicted}. Moreover, \emph{all} cases assigned with TBS $2$ and $6$ by the algorithm are indeed benign and malignant, respectively, demonstrating the potential of using the algorithm as a screening tool. Moreover, as can be seen in the figure, these predictions include cases which were considered indeterminate by the experts, \emph{i.e.}, with TBS categories $3$, $4$ and $5$. Namely, in these cases, the algorithm provides better decisions than human implying on the potential to use the algorithm as an assisting tool for pathologists in indeterminate cases.

\section{Conclusions} 
We have addressed the problem of thyroid-malignancy prediction from whole-slide images by developing an algorithm that mimics a pathologist, who identifies groups of follicular cells and categorizes the slides according to them, based on the Bethesda system. We have further introduced a framework for the simultaneous prediction of thyroid malignancy and TBS. Experimental results demonstrate that the proposed algorithm achieves performance comparable to cytopathologists. Moreover, through the prediction of the TBS categories, we have shown the potential of the algorithm to be used for screening as well as improving indeterminate prediction. In future study, we plan to address the computation challenge of exploring the full resolution of the scans at different focus values.  We further plan to significantly increase the number of the tested slides, the comparison to expert pathologists, and to address the detection of uninformative non-diagnostic slides.


\appendix

\section*{Appendix A.}
\paragraph{Networks} Both networks share the same architecture based on VGG11, which is presented in Table \ref{tab:vgg11}. The networks are trained using stochastic gradient descent with learning rate $0.001$, momentum $0.99$ and weight decay with decay parameter $10^{-7}$.

\begin{table}[htbp]
  \begin{centering}
  {\scriptsize{}}%
  \begin{tabular}{|c|c|}
\hline 
\multicolumn{2}{|c|}{\textbf{\scriptsize{}Feature extraction layers}}\tabularnewline
\hline 
\hline 
\textbf{\scriptsize{}Layer} & \textbf{\scriptsize{}Number of filters}\tabularnewline
\hline 
{\scriptsize{}conv2d} & $64$\tabularnewline
\hline 
{\scriptsize{}Max-pooling(M-P)} & \tabularnewline
\hline 
{\scriptsize{}conv2d} & $128$\tabularnewline
\hline 
M-P & \tabularnewline
\hline 
{\scriptsize{}conv2d} & $256$\tabularnewline
\hline 
{\scriptsize{}conv2d} & $256$\tabularnewline
\hline 
M-P & \tabularnewline
\hline 
{\scriptsize{}conv2d} & $512$\tabularnewline
\hline 
{\scriptsize{}conv2d} & $512$\tabularnewline
\hline 
M-P & \tabularnewline
\hline 
\end{tabular}{\scriptsize\par}
\par\end{centering}
\medskip{}

\begin{centering}
\begin{tabular}{|c|c|}
\hline 
\multicolumn{2}{|c|}{\textbf{\scriptsize{}Classification layers}}\tabularnewline
\hline 
\hline 
\textbf{\scriptsize{}Layer} & \textbf{\scriptsize{}Output size}\tabularnewline
\hline 
{\scriptsize{}Linear} & {\scriptsize{}$4096$}\tabularnewline
\hline 
{\scriptsize{}Linear} & {\scriptsize{}$4096$}\tabularnewline
\hline 
{\scriptsize{}Linear} & {\scriptsize{}$1$}\tabularnewline
\hline 
\end{tabular}{\scriptsize\par}
  \par\end{centering}  
  \caption{VGG11 based architecture used for both the first and the second neural
  networks in the proposed algorithm. Each conv2d layer comprises 2D
  convolutions with the parameters $\text{kernel\_size}=3$ and $\textrm{padding}=1$.
  Parameters of the Max-pooling layer: $\textrm{kernel\_size}=2$, $\textrm{stride}=2$.
  The conv2d and the linear layers (except the last one) are followed
  by batch normalization and ReLU.}
  \label{tab:vgg11}
\end{table}

\end{document}